\documentclass[conference]{IEEEtran}
\IEEEoverridecommandlockouts
\usepackage{times}

\usepackage[numbers]{natbib}
\usepackage{multicol}
\usepackage{amsmath}
\usepackage{array}
\usepackage[bookmarks=true]{hyperref}
\usepackage{graphicx}
\usepackage{xcolor}
\usepackage{capt-of} %

\newcommand{\sofar}[1]{#1}
\newcommand{\sysname}{HardwareRSI}
\newcommand{\benchname}{HardwareRSI-Bench}

\begin{document}

\title{Why Stop at Software? Recursive Self-Improvement\\Reaches Robot Hardware}

\author{%
  \IEEEauthorblockN{%
    Yida Yang\textsuperscript{1,2,3},\quad
    Ao Zhou\textsuperscript{1,2,3},\quad
    Wei He\textsuperscript{1,2,3},\quad
    Guang Chen\textsuperscript{1,2,3,*}%
  }
  \IEEEauthorblockA{%
    \textsuperscript{1}Physical Intelligence Lab, Tongji University, Shanghai, China\\
    \textsuperscript{2}Shanghai Innovation Institute, Shanghai, China\\
    \textsuperscript{3}Robotscale AI, Shanghai, China%
  }
  \thanks{%
    First author: Yida Yang
    (\href{mailto:yida@mail.tongji.edu.cn}{yida@mail.tongji.edu.cn}).\textsuperscript{*}Corresponding author: Guang Chen
    (\href{mailto:guangchen@tongji.edu.cn}{guangchen@tongji.edu.cn}).
  }%
}

\input{sections/teaser}
\maketitle

\begin{abstract}
Agentic robots can improve their control programs and manipulation skills through experience, yet the hardware through which they act and sense typically remains fixed.
We introduce \sysname{}, to our knowledge the first framework for recursive self-improvement (RSI) of robot hardware in which a closed-loop manipulation agent can redesign its own hardware, including its mechanisms and sensors.
The agent identifies limitations of its current hardware from execution feedback and uses observations from each modified design to guide the next hardware change.
We further develop \benchname{} to evaluate RSI of robot hardware across eight classes of simulated manipulation tasks drawn from real jobs. Within a budget of 30 attempts per task, the agent completes \sofar{33} of \sofar{36} tasks. It completes \sofar{3} with its original hardware and \sofar{30} after redesign, with multiple hardware changes on \sofar{21} of those \sofar{30} tasks.
To test whether one hardware design can serve a family of related tasks, we evaluate \sofar{four} families of three tasks each. In \sofar{three} of these families, the agent solves all three tasks with the same design. A single design also completes a four-stage assembly task spanning multiple classes.
The agent keeps improving after its first success: more efficient designs and practice cut the control steps a task takes by a median of \sofar{22\%}.
On a physical robot, the agent uses feedback from a failed box lift to design gripper sleeves and successfully lifts the box with them.

\end{abstract}

\IEEEpeerreviewmaketitle

\section{Introduction}
\label{sec:intro}

Recent work on recursive self-improvement (RSI) in robotics has shown that agents can use feedback from one attempt to refine their control programs and manipulation skills for the next~\cite{song2026embodiedrsi,tian2026physevo,liang2026robofoundry,lang2026embodirsi}.
EmbodiedRSI refines code and skills together, improving performance on tasks that combine familiar skills in ways not seen during pretraining~\cite{song2026embodiedrsi}. PhysEvo further adapts skills developed in simulation by diagnosing failures on a physical robot and testing corrections in subsequent trials~\cite{tian2026physevo}. In these systems, however, the robot’s hardware configuration, including its mechanisms and sensors, is chosen before it attempts the task and remains unchanged as its skills improve. Consider a robot tasked with retrieving a part from the bottom of a deep bin. If its gripper is too wide to pass through the opening, changing the grasp trajectory alone cannot bring the part within reach. Can an agent identify such a physical limitation from its own attempts, redesign the component responsible, and use feedback from trials with the modified hardware to decide what to change next?

One way to overcome such a physical limitation is to ask a design agent to develop a suitable tool or modify the robot itself. VLMgineer~\cite{gao2026vlmgineer} searches in simulation for a tool and an action sequence suited to the task. Its cube-retrieval prompt identifies the reach limitation and explicitly requests a tool before the robot makes an attempt. RobotSmith~\cite{lin2025robotsmith} jointly optimizes tool geometry and the trajectory for using the tool. Its design updates rely on rendered views of candidate tools and simulator scores, leaving the operating robot no role in deciding what hardware to change. RoboBuilder~\cite{nehme2026robobuilder} uses external video of a failed run to redesign a physical vehicle. The vehicle runs open loop, and the reported experiments end after a single round of redesign, without using evidence from the modified vehicle to guide a further change. These methods automate hardware design, but none demonstrates RSI by an operating agent that identifies a physical limitation through its own attempts and uses observations from the modified hardware to guide its next redesign.

We introduce \sysname{} (Fig.~\ref{fig:teaser}), to our knowledge the first framework for recursive self-improvement of robot hardware in which a manipulation agent can redesign both its own mechanisms and its sensing configuration based on experience from its own task attempts. The agent uses execution feedback to decide whether to revise its control program or change the hardware. It specifies hardware changes through CAD programs that define the geometry of individual parts and an assembly file that specifies how parts, sensors, and actuators are mounted. The same agent then resumes the task with the modified robot. A change in geometry can make new motions and contacts possible, while a change in sensor placement can reveal constraints that were previously unobservable. Each hardware design thus serves two roles: it embodies what the agent has learned from earlier attempts and determines what evidence the agent can gather to guide its next redesign. In the simulated task in Fig.~\ref{fig:teaser}(b), the agent repositions its wrist camera to inspect a failed suction grasp inside a deep bin. The new view reveals that the bracket’s flat surface is too narrow for the cup to seal, prompting the agent to replace the cup with long fingers. On a physical arm, the agent finds that its gripper can grasp only one corner of a wide box. It designs gripper sleeves and uses them to lift the box (Fig.~\ref{fig:teaser}(c)).

We develop \benchname{} to evaluate how manipulation agents use experience with their current hardware to identify physical limitations and guide subsequent redesigns. The benchmark comprises \sofar{36} simulated manipulation tasks drawn from real-world operations and organized into eight families by the demands they place on the robot’s hardware. Each task specifies an operational goal without identifying a limiting component or prescribing a modification. Constraints on reach, visibility, and contact arise from the geometry and physical interactions of the task itself. Across attempts, the agent retains its designs and execution history, allowing us to examine the observations that motivate each hardware change and what the agent can observe or accomplish with the resulting design. We test three hypotheses concerning the agent’s ability to overcome hardware limitations, reuse a single design across related tasks, and reduce the control steps required after initial success.

\textbf{H1. The agent can identify and overcome hardware limitations through its own trials.}
Within a budget of 30 attempts per task, the agent completes \sofar{33} of \sofar{36} tasks. It completes \sofar{3} with its original hardware and \sofar{30} after redesign. On \sofar{28} of the \sofar{30} tasks completed after redesign, trials with the original hardware reveal a specific limitation before the agent proposes a modification. On \sofar{21} of those \sofar{30} tasks, testing one modified design reveals a remaining limitation that guides the next hardware change.

\textbf{H2. One hardware design can serve multiple related tasks.}
We evaluate \sofar{four} task families, each comprising three related tasks that must be completed with the same hardware design. The control program can vary by task. The agent completes all three tasks with a single design in \sofar{three} of the \sofar{four} families. It also uses one design to complete all four stages of an electronic module assembly task, each drawn from a different family.

\textbf{H3. Refining hardware and control can reduce execution steps after the first success.}
For each of the \sofar{29} tasks completed more than once, we compare the number of control steps in the first and shortest successful runs. The median reduction is \sofar{22\%}. In a plate-handling task, the agent measures a flipper angle of \(76^\circ\) during execution and incorporates that angle into its next hardware design. Starting the flipper at the angle previously reached through motion reduces execution from 305 to 205 control steps.

\begin{figure*}[t]
\centering
\includegraphics[width=\textwidth]{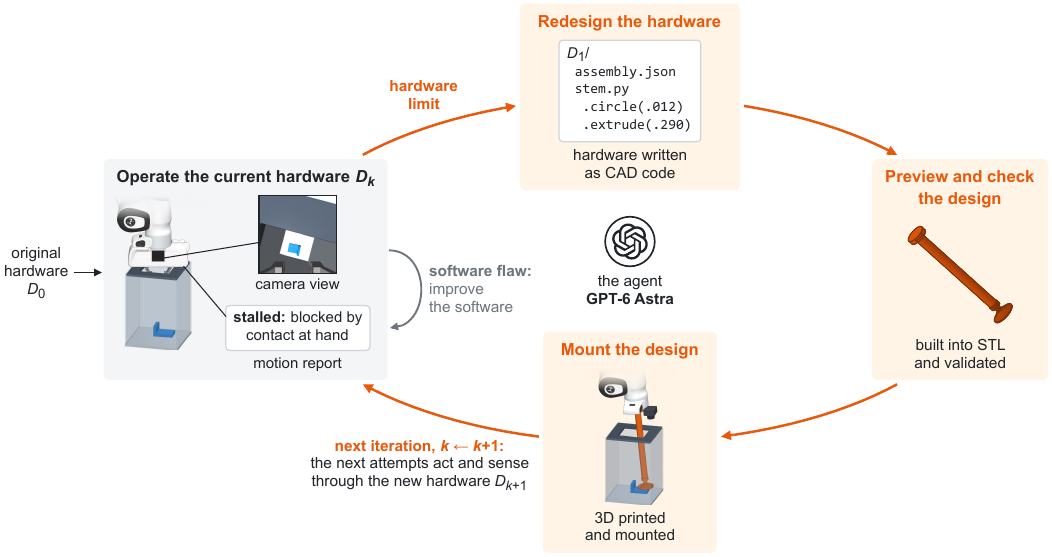}
\caption{\textbf{One iteration of \sysname{}.} One agent, here GPT-6 Astra (center), both operates the robot and redesigns its hardware. Each iteration has four stages (Sec.~\ref{sec:hardware-redesign}), illustrated with the bin task of Fig.~\ref{fig:teaser}(b) from the original hardware $D_0$ to $D_1$. Each motion with the current hardware $D_k$ returns images from every camera and the motion's final state; with $D_0$, the gripper is too wide to enter the bin and stalls on the rim. From this feedback, the agent decides whether a software flaw or a hardware limit is at fault. For a software flaw, it revises its control program and keeps the hardware (gray loop). For a hardware limit, it proposes the next design $D_{k+1}$ (orange), as in~(\ref{eq:hwrs-recursion}): CadQuery part programs and an \texttt{assembly.json} file, here for a 290\,mm rod that carries a suction cup. The builder exports the parts as STL meshes and checks the assembled design, which the agent can preview without spending an attempt. In simulation, mounting the design starts a new attempt; on the physical arm, the parts are printed from the same STL meshes and mounted. The next attempts act and sense through $D_{k+1}$, so the new design shapes the observations that guide the next redesign. With $D_1$, the rim hides the cup's contact, and moving the wrist camera in $D_2$ reveals why the cup cannot seal (Fig.~\ref{fig:teaser}(b)).}
\label{fig:system}
\end{figure*}

\section{Related Work}
\label{sec:related}

\subsection{Recursive Self-Improvement in Robotics}

Self-improving agents can rewrite their own code~\cite{zhang2025dgm,yin2025godel},
design agent workflows~\cite{hu2025adas}, and retain skills and memories
across tasks~\cite{wang2024voyager,ouyang2026reasoningbank}.
In robotics, language models have been used to select skills, generate
control programs, and correct failures from execution
feedback~\cite{ahn2022saycan,liang2023cap,liu2023reflect}.
Recent work on robotic RSI uses this experience to improve the software
that governs subsequent attempts~\cite{zhou2026whoimproves,luo2026rsiphysical}.
EmbodiedRSI tests proposed changes to code and skills through robot
trials~\cite{song2026embodiedrsi}. PhysEvo uses failed trajectories to
revise its skills and the software tools it uses to investigate
failures~\cite{tian2026physevo}. RoboFoundry updates memory and skills
from execution traces~\cite{liang2026robofoundry}, while EmbodiRSI uses
policy failures to guide data collection for subsequent policy
updates~\cite{lang2026embodirsi}. Yet in each of the robotic RSI
systems above, the hardware through which the robot acts and senses
remains fixed. To our knowledge, \sysname{} is the first framework for
RSI of robot hardware in which the agent controlling the robot can
redesign its hardware and change how its mechanisms and sensors
are mounted.

\subsection{Tool Design for Robotic Manipulation}

Robotic tool-design methods jointly optimize a tool and its use.
Liu et al.\ learn a policy that designs a tool for a specified goal
and a controller that uses it~\cite{liu2023tools}. PaperBot uses
physical trials to optimize how paper is cut into a
gripper~\cite{liu2024paperbot}. VLMgineer constructs tools from
rectangular blocks attached to the gripper fingers or flange and
specifies the motion waypoints before each simulated trial. In its
reach task, the prompt explicitly identifies the object as out of
reach and requests a tool~\cite{gao2026vlmgineer}. RobotSmith builds
a graspable tool from primitive or generated shapes. Its design
critic inspects renderings of the proposed tool before an optimizer
refines the tool geometry and trajectory in simulation~\cite{lin2025robotsmith}.
Both methods rely on privileged simulator information.
In VLMgineer's reach task, the environment code supplied to the model
contains the cube's dimensions, its initial and target positions,
and the reward function. RobotSmith supplies object geometry and poses
in the scene specification and scores candidates using task-specific
simulation metrics, such as the final position of the object in its
reach task~\cite{gao2026vlmgineer,lin2025robotsmith}.
These methods take the need for a tool as given. None requires the
agent operating the robot to discover from its own attempts whether
its hardware is the source of failure. In \sysname{}, the agent uses
its own observations to identify the limiting component and decide
whether to change its control program or redesign its hardware.

\subsection{Robot Co-Design and Reconfiguration}

Robot co-design jointly optimizes morphology and control. Most
methods optimize a robot for a single task, often locomotion,
and many use task-specific rewards to evaluate candidate designs
in simulation~\cite{wang2025codesignsurvey}. DERL evaluates 4,000
candidate morphologies per evolutionary run, using five million
simulated interactions to train the controller for each
morphology~\cite{gupta2021derl}. Transform2Act uses the same policy
to design and control the robot, but each episode starts from the
same initial morphology. Once control begins, the design remains
fixed for the rest of the episode~\cite{yuan2022transform2act}.
Auto-Robotist uses experience from earlier morphology searches to
guide its design agent in proposing new bodies~\cite{wang2026autorobotist}.
RoboBuilder incorporates physical failures into co-design. Given
the obstacle geometry and a calibrated simulator, a coding agent
designs a brick-built vehicle and its open-loop motor program.
The vehicle executes the program without sensor feedback. After
a failed physical run, the design agent reviews external video
in a new session for one round of redesign and
retesting~\cite{nehme2026robobuilder}. In these co-design systems,
execution feedback guides the search over candidate bodies.
Self-reconfigurable robots can change their own structure during
operation, but their changes are limited to joining, rearranging,
or shedding modules designed in
advance~\cite{yim2007modular,wyder2025metabolism,yu2026selfdestruction}.
These studies do not demonstrate recursive self-improvement of
robot hardware by the agent that operates it. In \sysname{}, the
agent redesigns its own hardware, attempts the task again, and
uses what it observes with the modified robot to guide its next
redesign.

\section{Methodology}
\label{sec:method}

In \sysname{}, the agent uses feedback from its own attempts to
decide whether to revise its control program or redesign the
robot's hardware. After a hardware change, the same agent resumes
the task with the modified robot. The new design shapes the
motions it can execute and the observations it can use to guide
a subsequent redesign (Fig.~\ref{fig:system}).

\subsection{Problem Setup and Task Objective}
\label{sec:problem-setup}

A language-model agent $\pi$ operates a robot with hardware design
$D$ and software $S$. In simulation, the initial design $D_0$
consists of a fixed-base, seven-degree-of-freedom Franka Panda arm,
a Franka Hand with two short fingers and an 80\,mm stroke, and a
wrist camera. The agent can attach parts and devices anywhere on the
arm or on components it has already added. It can also replace or remove
the hand, its fingers, or the wrist camera. The set $\mathcal{D}$
contains the designs that pass the construction and assembly checks
in Sec.~\ref{sec:hardware-redesign}. The software $S$ comprises the
agent's control scripts, skills, notes, and working context, which
persist across attempts.

A task $T$ specifies a workspace, a one-sentence instruction, and
a success condition. At the start of a session, the agent receives
the instruction, the available tools, and the operating rules. It
knows that the original hardware may be inadequate and must
determine which physical limitation, if any, matters for the task.
During execution, it receives images from a fixed overview camera
and the cameras on the robot,
readings from its sensors, and feedback on why each motion stops.
At the end of an attempt, the evaluator reports the task outcome
and the number of control steps. The agent has no direct access to
the simulator state used to evaluate success.

We write $\tau=A_T(D,S)$ for an attempt at task $T$ with hardware
$D$ and software $S$. Each simulated attempt starts from the same
prescribed workspace layout. It ends when the goal condition holds,
when a failure condition is met, or after $N=5{,}000$ control
steps of 50\,ms each. Only commands that run the robot consume
control steps. A single-task session allows up to $R=30$ attempts.
The objective is first to complete the task and then to reduce the
number of steps required for a successful execution. After its
first success, the agent may continue with the same hardware or
mount a new design until it chooses to stop or exhausts the budget.
Mounting a design starts a new attempt.

We test whether one hardware design can meet several task demands.
In the first, $m$ tasks from one family share a budget of $mR$
attempts. One hardware design must complete all $m$ tasks, although
the control program may vary by task. In the second, stages from
different families form a single task chain in one workspace.
The robot must complete every stage in one attempt with the same
design.

\subsection{Recursive Improvement of Hardware and Software}
\label{sec:recursive-update}

\textbf{Improving control with fixed hardware.}
Let $c(\tau)$ equal one for success and zero for failure, and let
$|\tau|$ count control steps. We rank attempts lexicographically
by $(c(\tau),-|\tau|)$, first by success and then, for the same
outcome, by fewer steps. For a fixed design $D\in\mathcal{D}$,
the best score achievable by changing the software is
\begin{equation}
    V_T(D)=\max_S \bigl(c(\tau),-|\tau|\bigr),
    \qquad \tau=A_T(D,S).
    \label{eq:hwrs-ceiling}
\end{equation}
Revising a control program or skill can improve the score attained
with $D$, while the ceiling $V_T(D)$ remains fixed%
~\cite{song2026embodiedrsi,tian2026physevo,liang2026robofoundry,lang2026embodirsi}.
A hardware change can raise the ceiling, making a previously
infeasible task feasible or reducing the number of steps needed to
complete it.

\textbf{Redesigning hardware from experience.}
Let $D_k$ denote the $k$th design mounted in a session and $S_k$
the software available when it is mounted. The agent may make
several attempts with $D_k$. The record $E_k$ contains the
commands issued, the observations returned during execution, and
the outcome and step count of each attempt.

To choose between revising control and changing hardware, the agent
examines the motions and contacts that lead to the observed outcome.
In a pin-insertion task, an image of the pin approaching
the hole off-center may prompt a correction to the commanded pose.
If a joint reaches its limit while the pin is still short of the
hole, the agent can instead examine whether moving the gripper
along its mount would extend its reach. When the agent chooses a
hardware change, it uses the records accumulated across designs to
propose the next design and its accompanying software,
\begin{equation}
    (D_{k+1},S_{k+1})=\pi(E_0,\ldots,E_k).
    \label{eq:hwrs-recursion}
\end{equation}
The agent $\pi$ proposes this change directly from its trial history,
without a task-specific design reward or a search over candidate
designs. It then operates $D_{k+1}$. The geometry of $D_{k+1}$
determines which motions and contacts are possible, and its sensors
determine which aspects of those interactions the agent can observe.
Thus $D_{k+1}$ is both the result of the update and the hardware
through which the agent acquires $E_{k+1}$ for a subsequent
redesign~\cite{zhou2026whoimproves}.

In the simulated bin task (Fig.~\ref{fig:teaser}(b)), a suction cup
reaches the bracket, but the bin rim obscures the contact. By moving
the wrist camera, the agent can see that the bracket's flat face
is too narrow for the cup to seal. That observation guides its
switch to long fingers. Experience can also inform geometry
directly. In a plate-handling task, the agent measures a useful
flipper angle of $76^\circ$ during execution and builds that angle
into its next design. The revised flipper starts at the orientation
the agent has previously reached through motion.

\subsection{Hardware Redesign from Execution Feedback}
\label{sec:hardware-redesign}

\textbf{Operating the current hardware.}
The agent operates the robot through 17 tools built on
Embodify~\cite{yang2026embodify} and exposed through the Model
Context Protocol~\cite{anthropic2024mcp}. It can invoke a tool
directly or write a script that sequences several
calls~\cite{liang2023cap}. Each motion returns images from every
camera and an explanation of why it stops. This feedback can
indicate a stall short of the target, a joint at its position or
torque limit, a singular configuration, or excessive contact force.
The agent can inspect these images alongside joint, actuator, and
sensor readings when interpreting the outcome of a motion.

\textbf{Redesigning the hardware.}
Each design consists of CadQuery~\cite{cadquery} part programs and
an \texttt{assembly.json} file. The programs define the geometry
of printable plastic or rubber parts. The assembly file specifies
the end effector and the mounting pose of each component relative
to the arm or another component. To lengthen a finger, the agent
edits its part program. To move a camera, it edits the camera's
mounting pose. A part can carry a free or spring-loaded
joint, or a joint driven by a servo or linear actuator. A component
catalog provides dimensions, masses, and operating limits for
grippers, suction cups, cameras, borescopes, and force, tactile,
and range sensors.

\textbf{Previewing and checking the design.}
The builder runs each part program in isolation, checks that the
resulting solid is closed and connected, verifies a minimum
thickness of 2\,mm and a maximum length of 1\,m, and exports an
STL mesh. It assembles the design in MuJoCo~\cite{todorov2012mujoco}
and checks mounting relationships, component overlap, and payload.
The simulation models collisions involving added parts, limits
actuator forces and torques to their ratings, and permits suction
only when the cup sits flat on a sealable surface. With
\texttt{preview\_design}, the agent can run these checks and
inspect the views from the proposed camera poses without spending
an attempt. The builder returns all detected issues and the
original CAD errors for correction. Passing the checks establishes
that the design can be loaded. Its effect on the task is tested
in the next attempt.

\textbf{Mounting the design and starting the next iteration.}
The agent calls \texttt{load\_design} with a note explaining what
has changed and why. In simulation, the call mounts the design and
starts a new attempt. A newly mounted camera returns images on
the next motion, and new sensors and actuators become available
through the robot interface. On the physical arm, parts are printed
from the same STL meshes and mounted before the agent resumes the
task (Sec.~\ref{sec:exp-real}).

Design files, execution logs, and working notes persist across
attempts. The agent can consult its notes when its conversation
history is compacted. For a family of tasks, the interface accepts
a task index, identifies each hardware version by its contents,
and records which tasks that version has completed. The agent can
therefore return to an earlier design and try it on an outstanding
task. After operating the modified robot, it decides from the
resulting observations whether to refine its control, redesign
the hardware again, or end the session.

\section{Experiments}
\label{sec:experiments}

\subsection{\benchname{} and Experimental Setup}
\label{sec:benchmark}
\label{sec:setup}

\begin{figure*}[t]
\centering
\includegraphics[width=\textwidth]{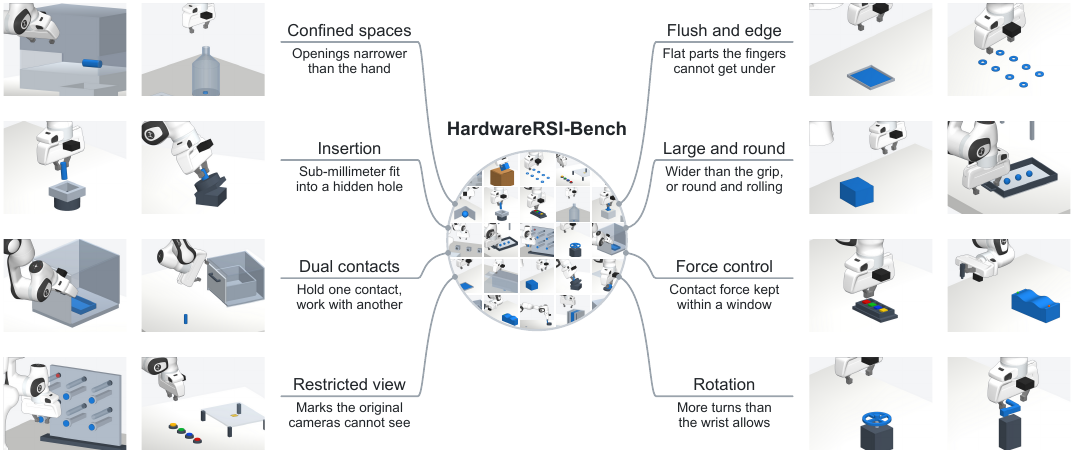}
\caption{\textbf{The eight task families in \benchname{}.}
Each family groups tasks by a physical demand they share. Two
tasks per family are shown with the original hardware. The center
shows additional benchmark scenes.}
\label{fig:bench}
\end{figure*}

\textbf{Tasks.}
We design \benchname{} to evaluate whether an agent can discover
a hardware limitation through its own attempts and redesign the
robot to overcome it. The benchmark comprises \sofar{36} simulated
manipulation tasks drawn from work in warehouses, workshops,
assembly lines, quality-control stations, and laboratories. Each
instruction specifies the operational goal, such as ``open the valve
handwheel four turns.'' The agent must determine which physical
constraints matter as it attempts the task.

The eight task families group operations by the demands they place
on the robot (Fig.~\ref{fig:bench}). In \emph{confined spaces}, the
robot must reach through openings narrower than its hand.
\emph{Insertion} tasks require fitting a pin into a hole with
submillimeter clearance. \emph{Dual-contact} tasks require
maintaining one contact while making another, as when holding a
carton flap open to place a box inside. \emph{Restricted-view}
tasks require inspecting features the original cameras cannot
see, such as a mark inside a blind hole.

\emph{Flush-and-edge} tasks provide little clearance for the
fingers to get under or around an object, as when extracting a
book from a packed shelf. \emph{Large-and-round} tasks involve
objects that exceed the gripper's 80\,mm span or roll against its
flat finger pads. \emph{Force-control} tasks require maintaining
contact force within a prescribed range, although the original
hardware lacks a force sensor. \emph{Rotation} tasks require
turning a part through multiple revolutions even though the wrist
cannot complete a full revolution in one motion.

\begin{figure*}[t]
\centering
\includegraphics[width=\textwidth]{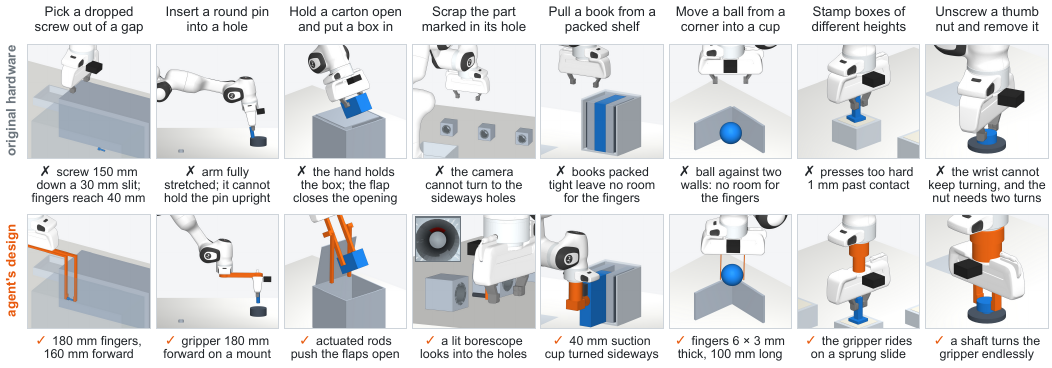}
\caption{\textbf{Hardware changes across the eight task families.}
Each column shows one task. The top row shows the original hardware
during an attempt and identifies the constraint it encounters.
The bottom row shows the design used in the shortest successful
attempt. Agent-designed parts are orange and task objects are blue.}
\label{fig:tasks}
\end{figure*}

Workspace dimensions, clearances, and friction take the values of the
real-world operations the tasks represent. A constraint may emerge only during
execution, for instance when the robot makes contact or its view
becomes occluded. The simulator checks a success
condition against the workspace state. In object-placement tasks,
for example, the entire object must reach its target and remain at
rest there for 0.5\,s. Figure~\ref{fig:tasks} shows the original
hardware and a successful redesign for one task from each family.

\textbf{Agent and protocol.}
\label{sec:exp-cost}
We run \sysname{} with GPT-6 Astra at high reasoning effort in the
Codex CLI. The simulated platform is a Franka Panda in
robosuite~\cite{zhu2020robosuite} and MuJoCo~\cite{todorov2012mujoco}.
Each task has one session of up to 30 attempts, starting with the
original hardware. The agent decides whether and when to change
its hardware. We count a task as completed if any attempt in the
session succeeds and record whether the original hardware
succeeds. The median session uses \sofar{6} attempts.

For H2, we select three tasks from each of \sofar{four} families.
The three tasks within a family share one session and a budget
of 90 attempts. The agent can use a different control program
for each task, but a single hardware design must complete all
three. We also test a four-stage electronic module assembly task.
Its stages draw on four different families and must be completed
in one workspace with the same design.

\subsection{Results}
\label{sec:results}

\begin{table}[t]
\centering
\caption{\textbf{Results by task family.}
Of \sofar{36} tasks, \sofar{3} are completed with the original
hardware and \sofar{30} after redesign, for \sofar{33} in total.
Step reduction is the median percentage decrease in control steps
from the first to the shortest successful attempt among tasks
completed at least twice. Shared design reports whether one version
completes the three tasks selected from a family for H2. A dash
marks families outside that evaluation.}
\label{tab:main}
\footnotesize
\setlength{\tabcolsep}{3.6pt}
\begin{tabular}{@{}lrrrrc@{}}
\hline
 & & \multicolumn{2}{c}{Tasks completed} & & \\
\cline{3-4}
Family & Tasks & Original & Any design &
\shortstack{Step\\reduction} & \shortstack{Shared\\design} \\
\hline
Confined spaces & \sofar{5} & \sofar{0} & \sofar{5} & \sofar{13\%} & \sofar{yes (v19)} \\
Insertion       & \sofar{5} & \sofar{1} & \sofar{5} & \sofar{16\%} & \sofar{yes (v1)} \\
Dual contacts   & \sofar{5} & \sofar{0} & \sofar{4} & \sofar{18\%} & \sofar{no} \\
Restricted view & \sofar{5} & \sofar{1} & \sofar{5} & \sofar{62\%} & \sofar{yes (v2)} \\
Flush and edge  & \sofar{4} & \sofar{0} & \sofar{4} & \sofar{25\%} & -- \\
Large and round & \sofar{4} & \sofar{1} & \sofar{4} & \sofar{19\%} & -- \\
Force control   & \sofar{4} & \sofar{0} & \sofar{2} & \sofar{27\%} & -- \\
Rotation        & \sofar{4} & \sofar{0} & \sofar{4} & \sofar{31\%} & -- \\
\hline
All & \sofar{36} & \sofar{3} & \sofar{33} & \sofar{22\%} & \sofar{3/4} \\
\hline
\end{tabular}
\end{table}

\textbf{H1. The agent can identify and overcome hardware limitations
through its own trials.}
\label{sec:exp-h1}
The agent completes \sofar{33} of \sofar{36} tasks within the attempt
budget. It completes \sofar{3} with the original hardware and
\sofar{30} after redesign, with at least one task completed after
redesign in every family (Table~\ref{tab:main}). Among completed
tasks, the median is \sofar{four} attempts to the first success.

For all \sofar{30} tasks completed after redesign, the agent
records a specific limitation before proposing its first hardware
change. In \sofar{28} of these tasks, it identifies the limitation
by trying the original hardware. The resulting designs reflect
the constraints encountered during execution (Fig.~\ref{fig:tasks}).
Among tasks completed after redesign, every restricted-view task
uses a borescope, every dual-contact task uses a linear actuator,
and every rotation task uses a servo in multi-turn or continuous
mode. Both successful force-control tasks use a spring-loaded
slide. In all, \sofar{23} of the \sofar{30} tasks are completed
with a design that incorporates a repositioned camera, a sensor,
an actuator, a spring, or a suction cup.

Operating a modified design can expose a further limitation.
In the crank task, a multi-turn servo with an offset socket drives
the wrist to its angular limit. The agent turns the servo sideways
in its second design and adds a camera in its third to observe
the socket engaging the crank. With a square socket on its fourth
design, it turns the crank five times. Across \sofar{21} of the
\sofar{30} tasks completed after redesign, feedback from operating
one modified design guides a subsequent hardware change.

\textbf{H2. One hardware design can serve multiple related tasks.}
\label{sec:exp-h2}
In \sofar{three} of the \sofar{four} families evaluated, one design
completes all three selected tasks in that family
(Table~\ref{tab:main}). For insertion, the first design completes
all three by attempt 10. For restricted view, the second design
does so by attempt 14. In confined spaces, earlier designs complete
different subsets of the tasks before the nineteenth design
completes all three by attempt 45.

The electronic module assembly task tests whether one design can
also meet different hardware demands over successive stages of a
single operation. On attempt 12, the eighth design completes all
four stages without a hardware change between stages.

\begin{figure}[t]
\centering
\includegraphics[width=\columnwidth]{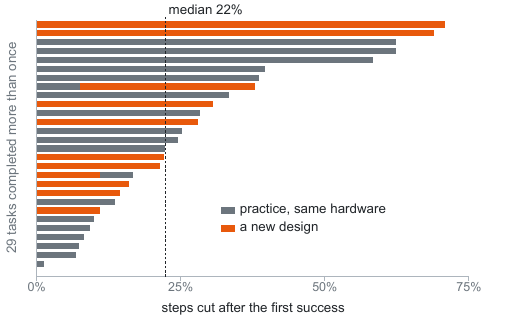}
\caption{\textbf{Control steps saved after the first success.}
Each bar represents one of the \sofar{29} tasks completed at least
twice. Its length is the percentage reduction in control steps from
the first to the shortest successful attempt. Gray denotes reductions
observed with the same hardware and orange denotes reductions
observed after a hardware change. Both can occur within a task.
The dashed line marks the median across all \sofar{29} tasks
(\sofar{22\%}).}
\label{fig:steps}
\end{figure}

\textbf{H3. Refining hardware and control can reduce execution steps
after the first success.}
\label{sec:exp-h3}
Of the \sofar{29} tasks completed more than once, \sofar{28} have a
later successful attempt that uses fewer control steps than the
first. Across all \sofar{29} tasks, the median reduction from the
first to the shortest successful attempt is \sofar{22\%}
(Fig.~\ref{fig:steps}).

Shorter successful executions occur with unchanged hardware on
\sofar{19} tasks and after a hardware change on \sofar{11}. Some
tasks show both patterns. In the tray-carrying task, the agent
refines its control program with the same hardware and reduces
the execution from 462 to 174 steps. In the crank task, a later
design with a locating sleeve completes the task in 1,101 steps,
down from 3,551.

In the plate-handling task, a measurement made during execution
informs the geometry of the next design. The agent
measures a useful flipper angle of $76^\circ$ while operating its
second design and builds that angle into the third. The flipper
then starts at an orientation the robot has previously reached
through motion, shortening the sequence from 305 to 205 control
steps.

\subsection{Evaluation on a Physical Robot}
\label{sec:exp-real}

We evaluate \sysname{} across repeated trials on a physical robot
(Fig.~\ref{fig:teaser}(c)). The agent controls a seven-degree-of-freedom
AgileX NERO arm equipped with a Pika gripper and a wrist-mounted Intel
RealSense D405 depth camera. The original gripper opens to approximately
90\,mm. Asked only to lift the box, the agent chooses a corner grasp for
its first attempt. The box slips during the lift.

From where its fingers have made contact and the slip it has observed, the
agent infers that its gripper cannot span the box. It designs a
pair of sleeves that fit over the original fingers to increase their
effective span, specifying the sleeve geometry in CadQuery and the
mounting positions in the assembly file. Once the sleeves are printed
in PLA and mounted, the same agent resumes the task and lifts the box.
With the same sleeves, it reduces the lift time from 15.6\,s on its
first successful attempt to 9.0\,s on a later trial. These physical
trials support H1 by showing that the agent can discover a hardware
limitation through its own attempts and overcome it with a gripper
it has redesigned.

\section{Conclusion}
\label{sec:conclusion}
To our knowledge, \sysname{} is the first framework for recursive
self-improvement of robot hardware. \sysname{} extends recursive self-improvement (RSI) to the hardware
through which a robot acts and senses. The same agent that encounters
a physical limitation can revise its control program or change the
hardware it will use on its next attempt. Of the 30 simulated tasks
solved after hardware redesign, 21 involve more than one revision.
A single design can serve several related tasks, and experience
gained after an initial success can guide changes that shorten
execution. On a physical arm, feedback from a failed grasp guides
the design of gripper sleeves that enable the robot to lift a box
wider than its original gripper can span. A hardware revision can contribute to improvement even when the next
attempt still fails. A change in geometry can make a new contact
possible, while a change in sensing can reveal why an action failed.
By operating the modified robot, the agent can obtain evidence that
was unavailable with the previous design and use it to decide what
to change next. In this way, the robot's hardware shapes the process
by which the agent improves that hardware through experience.

\bibliographystyle{plainnat}
\bibliography{references}

\end{document}